18th CIRP Conference on Intelligent Computation in Manufacturing Engineering

# Physics-Informed Machine Learning for Smart Additive Manufacturing

Rahul Sharma[a,b], Maziar Raissi[c], Y.B. Guo[a,b]*

*[a] Dept. of Mechanical and Aerospace Engineering, Rutgers University-New Brunswick, Piscataway, NJ 08854, USA*
*[b] New Jersey Advanced Manufacturing Institute, Rutgers University-New Brunswick, Piscataway, NJ 08854, USA*
*[c] Dept. of Mathematics, University of California, Riverside, CA 92521, USA*

* Corresponding author. Tel.: +1-848-445-2225; fax: +1-732-445-3124. *E-mail address:* yuebin.guo@rutgers.edu

## Abstract

Compared to physics-based computational manufacturing, data-driven models such as machine learning (ML) are alternative approaches to achieve smart manufacturing. However, the data-driven ML's "black box" nature has presented a challenge to interpreting its outcomes. On the other hand, governing physical laws are not effectively utilized to develop data-efficient ML algorithms. To leverage the advantages of ML and physical laws of advanced manufacturing, this paper focuses on the development of a physics-informed machine learning (PIML) model by integrating neural networks and physical laws to improve model accuracy, transparency, and generalization with case studies in laser metal deposition (LMD).

## 1. Introduction

The recent development in artificial intelligence (AI) and machine learning (ML) technologies demonstrates significant promise for advancements in the modeling of manufacturing processes. The effectiveness of traditional data-driven ML tools in data science is mainly due to the availability of vast amounts of labeled datasets, which can be acquired through experiments or physics-based simulations. However, data ingestion is very expensive and often restricts the feasibility of data-driven ML models as surrogate models for manufacturing processes.

Recently, the emergence of scientific machine learning, especially physics-informed machine learning (PIML), has introduced a new paradigm that leverages deep learning for solving partial differential equations (PDE) (1). This advancement enables the replacement of traditional numerical discretization methods, e.g., finite difference method (FDM) and finite element method (FEM), with neural networks (NNs) that can approximate PDE solutions. This framework integrates data with fundamental physical principles, such as the conservation laws of momentum, mass, and energy in laser powder bed fusion (LPBF) and laser direct energy deposition (DED), directly into the neural network architecture to guide a learning process. PIML offers an alternative to mesh-dependent methods like FDM and FEM, providing a mesh-free approach through automatic differentiation (2). This innovative strategy has the potential to navigate the complexities associated with the curse of dimensionality. Differing from traditional deep learning frameworks, PIML models can often be developed without the necessity for training data. Furthermore, with the advent of GPUs, these models can be trained significantly faster than conventional physics-based simulations.

On the side of growing advanced manufacturing technologies, metal additive manufacturing (AM) has been widely popular due to its unique ability to fabricate components with intricate shapes directly from digital blueprints. The market for metal AM has experienced substantial growth in the past two decades. Nonetheless, its integration into the manufacturing industry has not lived up to expectations. This is primarily due to the insufficient understanding of the reliable relationships between processes, microstructures, and properties. Since, metal AM process performance depends on its temperature gradient, therefore, understanding the transient thermal behavior of the process is very important to make an AM process smart.

Computational fluid dynamics (CFD) and finite element methods (FEM) models have been developed to understand the thermal behavior of AM processes. CFD model incorporating the phase change, latent heat, Marangoni flow, and recoil pressure due to the vaporization of material has been studied (3, 4). FEM models have been used to understand the macroscale phenomenon like thermal residual stress, distortion, and microstructure (5,6). Even though these models are comprehensive, they are so computationally expensive that they cannot be used for in-situ monitoring of the AM processes, which has limited their potential to enable a smart AM process.

In recent years, ML models have gained the attention of many researchers. Mozaffar et al. (7) developed a GRU-based stacked RNN model to predict the thermal history, assessing the effects of changes in geometry, build dimensions, tool-path strategy, laser power, and scan speed with promising accuracy in DED builds. Roy et al. (8) developed a surrogate model to predict the thermal history in the AM process. The model used the physics-based simulation data to train the surrogate model. Also, their approach leverages the G-Code directly instead of using the part geometry. The model shows a high predictive power at a low computational cost. Ren et al. (9) developed a combined Recurrent Neural Network and Deep Neural Network to predict the thermal field for laser-aided additive manufacturing (LAAM). Again, the training data was generated from a physics-based numerical model. The model can predict the thermal history of 100 different one-layer deposition cases having six different scanning strategies with good accuracy. Sarkar et al. (10) used the thermal video data from an infrared camera to train the ML model to study the thermal history of the submerged arc welding process. The model provides promising results with an accuracy of more than 93% for the prediction of maximum temperature in the domain. While ML models can process vast data sets for online monitoring purposes, such as predicting the temperature, they predominantly operate as "black-box" entities. Also, these models are very data-hungry as all these studies require physics-based simulations or massive experimental data to train the surrogate models.

Raisi et al. (11) developed a new physics-informed neural network (PINN) technique that integrates the underlying physical laws within the neural networks to solve the governing partial differential equations (PDEs) efficiently. Afterward, other researchers used this technique in different fields including manufacturing (12,13), but very few discussed training the PINN models without using any labeled training data. The capability of PINN to accurately predict the velocity and pressure field within the melt pool by using only the temperature training data has been investigated (14). In this study, a PINN model has been developed to predict the 3D temperature field in the LMD process. Owed to the nature of the PINN, the neural network is trained without using any labelled training data. Unlike traditional numerical methods that rely on meshes, PIML offers an approach that does not depend on meshes. The following sections of the paper are organized as follows: Section 2 describes the LMD process and the underlying physical laws, i.e., governing equations. Section 3 describes the architecture of the PIML model. Section 4 and Section 5 describe the results and conclusions respectively.

## Nomenclature

| | |
|---|---|
| $\rho$ | density of the material |
| $\kappa$ | thermal conductivity of the material |
| $\eta$ | laser absorptivity |
| $\sigma$ | Stefan-Boltzmann constant |
| $\varepsilon$ | emissivity of the material |
| $C_p$ | heat capacity of the material |
| $T$ | temperature |
| $q$ | heat flux |
| $r$ | laser beam radius |
| $d$ | distance from the laser center |

## 2. Problem definition

This work focuses on the deposition of a single layer of Ti-6Al-4V powders over the substrate of the same material in the LMD process. Fig. 1 shows the LMD process where metal powder is blown into the focused laser beam as it scans across a surface. Initially, the substrate is at room temperature. It is then scanned by a laser heat source that forms a small melt pool in which powder particles are converted into liquid and then solidified as the laser heat source moves to the next location. There are three different modes of heat transfer, i.e., conduction, convection, and radiation. To demonstrate the PIML model's ability to predict the thermal history of the LMD process, certain basic assumptions were made to simplify the model and ensure that the representability of the actual physical process is not compromised. These assumptions are given below:

- The deposited metal layer is thin compared to the substrate thickness; therefore, it is assumed that the deposited mass does not influence the macroscopic temperature.
- The latent heat of fusion is neglected.
- Melt pool fluid flow and evaporation are not considered.
- The substrate and deposited material are assumed to be homogeneous materials with constant values of the material properties.
- The laser heat source is assumed to have Gaussian heat distribution.
- Only half of the domain is analyzed to save the computation time because of symmetry along the x-z plane.

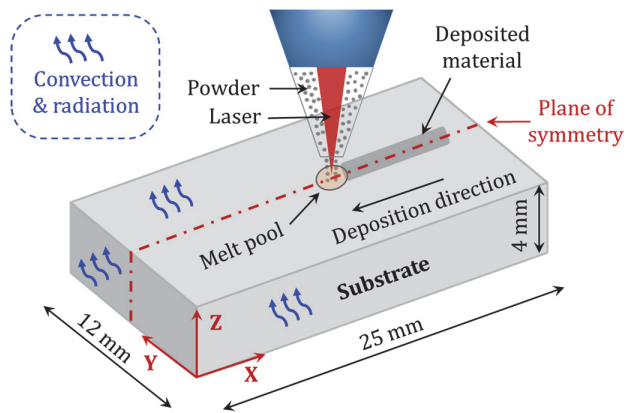

Fig. 1. Schematic of laser metal deposition (LMD).

### 2.1. Governing equations

The energy equation for LMD can be defined as (15):

$$\frac{\partial(\rho C_p T)}{\partial t} = \kappa \left( \frac{\partial^2 T}{\partial x^2} + \frac{\partial^2 T}{\partial y^2} + \frac{\partial^2 T}{\partial z^2} \right) \tag{1}$$

where $\rho$ is the density of the material, $C_p$ is the heat capacity, $T$ is the temperature and $\kappa$ is the thermal conductivity. The thermal boundary conditions are given by:

$$-\kappa \frac{\partial T}{\partial n} = Q_{laser} + Q_{conv} + Q_{rad} \tag{2}$$

where $n$ is the normal to the surface, $Q_{laser}$ is the heat input by the laser heat source, $Q_{conv}$ is the convective heat loss, and $Q_{rad}$ is the radiative heat loss and given by:

$$Q_{laser} = -\frac{2\eta P}{\pi r_b^2} \exp\left( \frac{-2d^2}{r_b^2} \right) \tag{3}$$

$$Q_{conv} = h\,(T - T_0) \tag{4}$$

$$Q_{rad} = \sigma \epsilon \,(T^4 - T_0{}^4) \tag{5}$$

where $\eta$ is the absorption coefficient, $P$ is the laser power, $r_b$ is the beam radius, $d$ is the distance from the laser center, $h$ is the convective heat transfer coefficient, $\sigma$ is the Stefan-Boltzmann constant, $\epsilon$ is the emissivity, and $T_0$ is the ambient temperature. The bottom surface of the substrate has a Dirichlet boundary condition with a fixed temperature equal to the ambient (298 K). Also, the X-Z plane has the symmetry boundary condition. The initial temperature of the domain is 298 K. The process parameters and material properties used in this study are tabulated below:

Table 1. Process parameters.

| Parameter | Value |
|---|---|
| Laser power (W) | 500 |
| Laser absorption coefficient | 0.4 |
| Laser beam radius (mm) | 1.5 |
| Laser scanning speed (mm/s) | 5 |

Table 2. Material properties of Ti-6Al-4V (16).

| Parameter | Value |
|---|---|
| Density (kg/m³) | 4122 |
| Heat capacity (J/kgK) | 831 |
| Thermal conductivity (W/mK) | 35 |
| Emissivity | 0.4 |

A validation dataset was generated using the FE-based COMSOL Multiphysics software. As shown in Fig. 2, the tetrahedron mesh with more element density on the top surface (due to a high thermal gradient) is used. There are a total of 8778 elements in the domain of size 25 mm × 6 mm × 4mm (half domain). A 3-second process was simulated. The temperature data with a 10 Hz frequency was extracted for the validation dataset.

### 2.2. PIML framework for thermal history

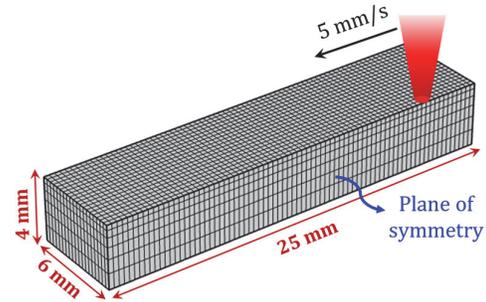

Fig. 2. Meshing of the numerical model.

Figure 3 shows the architecture of the PIML model used in this study. The basic nature of PIML is to use the physical governing laws in the form of PDEs residual, initial condition loss, boundary condition loss, and sometimes training data loss as well. It is worth noting that in the current study temperature data is not provided to train the model, hence no training data loss was used. The output (z) of the $n^{th}$ layer in the NN is given by:

$$z_n = \alpha_n(w_n^T z_{n-1} + b_n) \tag{6}$$

where $\alpha$ is the swish activation function (12), $w$ is the weight and $b$ is the bias. The network takes spatiotemporal coordinates (x, y, z, t) at the collocation points as the input and predicts the temperature value on the same points. These collocation points are similar to the mesh shown in Fig. 2 which has non-uniform distribution having more density towards the top surface due to the presence of a high thermal gradient on the top boundary. The model contains 4 hidden layers with 32 neurons per layer.

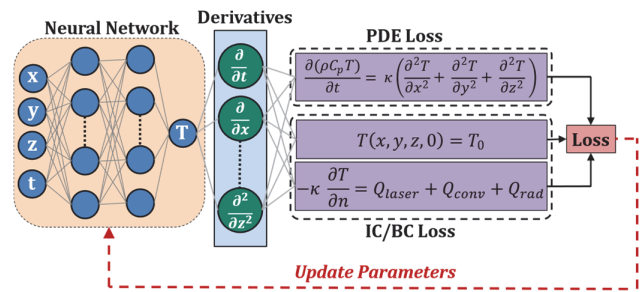

Fig. 3. PIML model used for the LMD process.

PyTorch was used to implement the model using Nvidia RTX A6000 GPU for model training. The model underwent 35,000 iterations where each set of 100 iterations took approximately 13 seconds. The network parameters are randomly initialized by the Glorot method. The model is trained using the Adam optimizer and a learning rate of 2e-4. Finally, the performance of the trained PINN model was compared to the benchmark FEA model by comparing the results at time = 1, 2, and 3 seconds.

The loss function of PIML makes it unique from conventional neural networks. The different losses used in the current study are given by:

$$\mathcal{L}_{PDE} = \frac{\partial(\rho C_p T)}{\partial t} - \kappa \left( \frac{\partial^2 T}{\partial x^2} + \frac{\partial^2 T}{\partial y^2} + \frac{\partial^2 T}{\partial z^2} \right) \quad (7)$$

$$\mathcal{L}_{IC} = T(x, y, z, 0) - T_0 \quad (8)$$

$$\mathcal{L}_{BC} = -\kappa \frac{\partial T}{\partial n} - Q_{laser} - Q_{conv} - Q_{rad} \quad (9)$$

where $\mathcal{L}_{PDE}$, $\mathcal{L}_{IC}$, and $\mathcal{L}_{BC}$ are PDE residual, initial condition loss, and boundary condition loss respectively. The total loss can be given as:

$$\mathcal{L}_{total} = (w_1 . \mathcal{L}_{PDE} + w_2 . \mathcal{L}_{IC} + w_3 . \mathcal{L}_{BC})/3 \quad (10)$$

where $w_1$, $w_2$, and $w_3$ are the weights of each source term. In this study, {1, 1e-4, 1} values are used for $w_1$, $w_2$, and $w_3$ respectively.

## 3. Results and discussion

To demonstrate the capability of the PIML model, the temperature evolution for single-layer deposition in the LMD process is studied. No labeled training data was used to train the PINN model. To quantify the prediction capabilities of the PINN, a validation dataset from COMSOL Multiphysics software was used. The model took approximately 1.26 hours to get trained. In the following subsections, the predicted temperature field comparison with validation data from FEA, the evolution of different losses during the training, and the temporal evolution of temperature during the heating and cooling stages are discussed.

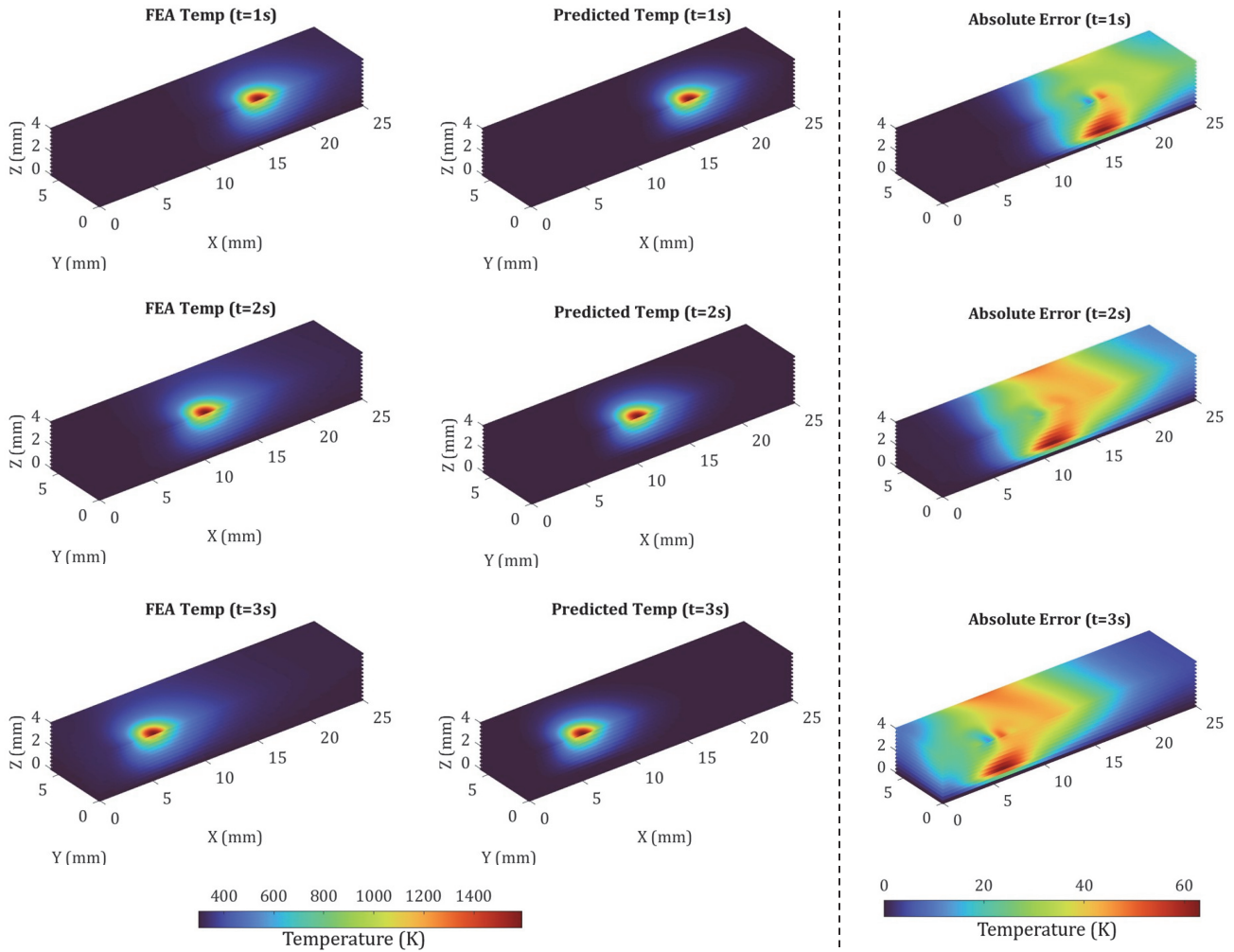

Fig. 4. Comparison of PIML-predicted temperature fields with FEA results and error distributions at different time steps.

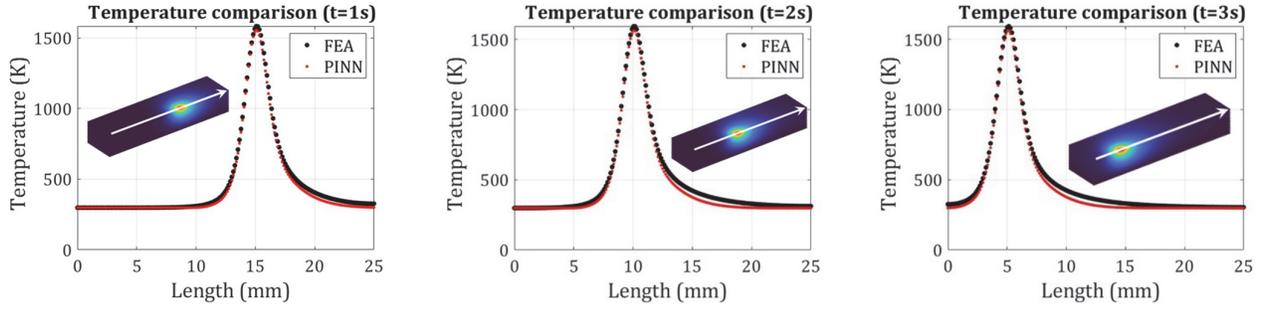

Fig. 5. Temperature distribution comparison along laser scanning path at different time steps.

### 3.1. Temperature field evolution during scanning

Figure 4 compares the predicted temperatures from the PIML model to the FEM-predicted ones and additionally displays the absolute errors in the prediction. The predicted temperature results match very well with the FEA data. The maximum temperature in the domain varies between 1550-1620 K for the whole metal deposition period. It exists on the top boundary due to constant heat input from the laser heat source. There is a steep temperature gradient around the laser center at every time instant. The maximum absolute error (MAE) in temperature calculation reported by the PIML model is 61.2 K with a 3.7% relative error. It is worth noting that the maximum absolute error location does not occur at the top surface. Rather, in the melt pool region on the top boundary MAE is approximately 34.7 K which comes to 2.1% for the relative error. This shows the capability of the PIML model to predict the temperature evolution without using any training data. Figure 5 shows the comparison of PINN predicted temperature with the FEA results on the top surface along the laser scanning path. It shows very good agreement that depicts the ability of PIML to accurately predict the melt pool length and width during the virtual monitoring. Even though this is beyond the scope of this study, this aspect will constitute the future research direction.

### 3.2. Model training efficiency

Figure 6 represents the different loss function evolutions during the model training. In the initial 6,000 epochs, all losses exhibit significant variability when using the Adam optimizer, yet they decline more rapidly and steadily in subsequent epochs under the L-BFGS optimizer. The total loss value is of the order $1e^{-2}$ which is acceptable for the current problem. Initially, the PDE residual loss is minimal because Equation (1) can be easily satisfied by a uniform temperature field, although the boundary conditions (BCs) outlined in Equations (2) to (5) are not satisfied at this point. As training progresses, the neural network incrementally adapts to the governing equations and boundary conditions of the system, aiming to equalize the magnitudes of the loss terms. It is worth noting that the initial condition loss does not converge to a very low value therefore, less weight is given to it in the total loss calculation (Eq. 10).

### 3.3. Heating and cooling predictions

Any location near the laser heat source experiences a heating and cooling cycle during the LMD process. The steep temperature gradient results in high thermal residual stresses. Sometimes, it results in the deformation of the material.

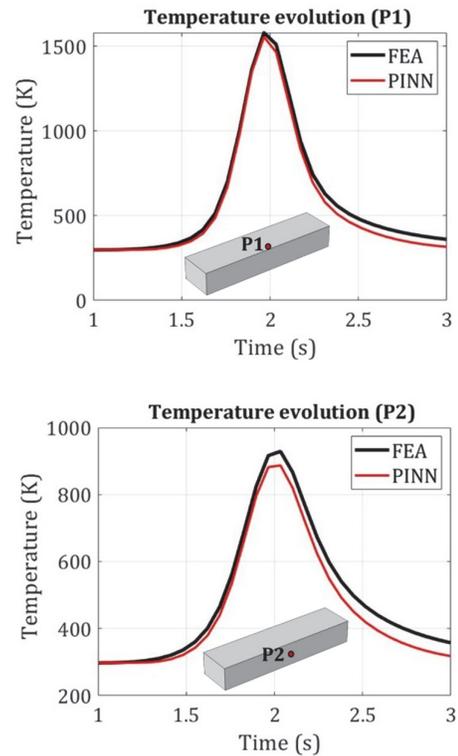

Fig. 7. Temporal evolution of temperature during heating and cooling.

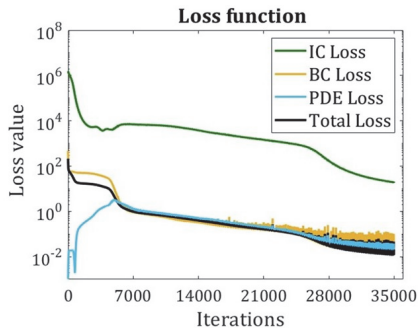

Fig. 6. Evolution of loss functions during model training.

Therefore, it is important to capture the correct temperature evolution at a particular location in the domain. Figure 7 shows the temporal evolution of temperature at two different locations during the material deposition (heating) and cooling stages. The first point (P1) is situated on the top surface of the laser scanning path while the second point (P2) lies 1.5 mm below the top surface. It can help in determining whether the PIML model can predict the correct length, width, and depth of the melt pool. It can be observed from the Fig. 7 that the PIML model predicts very well on the top surface while along the depth there is a minor deviation from the exact temperature from the FEA. This can be due to the fact that boundary conditions are explicitly incorporated in the loss function of PIML that helps to predict more accurately on the top surface than inside of the domain.

## 4. Conclusions and future work

The main challenge for the current data-driven ML models of smart manufacturing is the difficult acquisition of massive training datasets, which hinders the application of these methods in real manufacturing settings. This study develops a PIML model for the LMD process without using any labeled training data. The model is designed to predict full-field temperature histories. The predicted results are compared with the FEA results and the key conclusions can be enumerated:

- Prediction of temperature fields through PIML without any labeled training data is feasible and can be achieved by incorporating the loss of governing PDEs, initial conditions, and boundary conditions to the loss function.

- The PDE loss initially rises from a low to a high value before starting to decrease again after 6000 iterations. This occurs because the model can easily satisfy the governing equation with a uniform temperature value, but as the laser introduces more heat, the predictions become less accurate. Eventually, as the model adjusts its weights and biases, the PDE loss begins to decline once more.

- The PIML model predicts temperature very well on the top boundary with a relative error of 2.1% than inside the domain near the bottom with a relative error of 3.7%. This can be attributed to the explicit incorporation of boundary conditions in the loss function.

For outlook, additional physical laws may be integrated into the PINN model to predict multi-physical phenomena such as residual stress. The model's effectiveness in transfer learning may be evaluated when there are changes in process parameters and eventually act as a digital twin of an LMD process.

## Acknowledgments

The authors would like to thank the financial support of the National Science Foundation under the grant CMMI-2152908.

## Data availability

Data will be made available on request.